\documentclass[11pt]{article}

\usepackage[preprint]{acl}

\usepackage{times}
\usepackage{latexsym}
\usepackage{amsmath}
\usepackage{amssymb}
\usepackage{booktabs}

\usepackage{graphicx}
\usepackage{subcaption}
\usepackage{enumitem}
\usepackage{wrapfig}

\usepackage[T1]{fontenc}

\usepackage[utf8]{inputenc}

\usepackage{microtype}

\usepackage{inconsolata}

\usepackage{graphicx}

\title{What Breaks Under Pruning in Smart Homes, and When? Evaluating LLM Degradation Across Architectures and Task Complexity}

 \author{
    Congjing Zhang\thanks{Work done during an internship at Amazon.} \\
    Alexa Home AI, Amazon.com \\
    University of Washington \\
    \texttt{cjzhang@amazon.com} \\
    \texttt{congjing@uw.edu}
    \And
    Vashishtha Patil\thanks{Corresponding author.} \\
    Alexa Home AI, Amazon.com \\
    \texttt{pativash@amazon.com}
    \AND
    Henning Lange \\
    Alexa Home AI, Amazon.com \\
    \texttt{helange@amazon.com}
    \And
    Usman Aleem \\
    Alexa Home AI, Amazon.com \\
    \texttt{ualeem@amazon.com}
  }

\begin{document}
\maketitle
\begin{abstract}
Pruning can reduce the deployment cost of large language models (LLMs), but its impact on context-grounded tool calling remains poorly understood. We systematically study pruning-induced degradation in smart-home tool calling across four LLMs spanning dense Transformer, dense hybrid, and mixture-of-experts (MoE) architectures, together with depth, width, hybrid, and expert pruning methods. After post-pruning supervised fine-tuning (SFT), we evaluate more than 19,500 instances from three smart-home datasets. Beyond aggregate task accuracy, we characterize degradation along two dimensions: action components (i.e., operation, device, argument, and value) and task complexity. Our results show that dense models have narrow safe pruning regions followed by sharp degradation, while MoE models tolerate substantially more pruning. Pruning degrades grounded specificity before schema-level intent, and aggressive dense pruning can induce systematic over-refusal. These findings highlight the importance of evaluating pruning beyond aggregate accuracy when selecting pruned LLMs for reliable tool execution.
\end{abstract}

\section{Introduction}

Large language models (LLMs) are increasingly used as natural-language
interfaces for smart-home control, translating user requests into structured
actions grounded in household devices, capabilities, and states
\cite{rivkin2024aiot, zhang2026vcu}. However, deploying LLMs can incur
substantial memory and computational costs. Structured pruning reduces these
costs by removing components such as Transformer layers, channels, or experts
while preserving hardware-friendly model structures
\cite{ma2023llm, gao2024disp}. For deployment, an important question is
therefore not only how much an LLM can be pruned, but what capabilities are
lost as capacity is removed.

This question is particularly important for smart-home tool execution, where
a single action requires multiple decisions: what operation to perform, which
device to target, which argument to control, and what value to use
\cite{seo2026simuhome}. Requests also vary in complexity, from single-device
operations to multi-action, contextual, partially executable, and infeasible
requests \cite{li2025homebench}. Aggregate accuracy can therefore hide
substantially different pruning-induced failure modes. Prior work shows that pruning can disproportionately impair difficult
downstream tasks \cite{pmlr-v235-yin24b}, but existing studies largely
evaluate pruned LLMs using aggregate benchmarks, while smart-home benchmarks
primarily study unpruned models. It remains unclear how pruning sensitivity
varies across architectures, action components, and task complexity, or
whether aggressive pruning can qualitatively change model behavior rather
than simply reduce accuracy.

We systematically study these questions across four LLMs spanning dense
Transformer, dense hybrid, and sparse mixture-of-experts (MoE) architectures,
using representative depth, width, hybrid, and expert-pruning methods. Each
pruned model undergoes supervised fine-tuning (SFT), followed by evaluation
on more than 19,500 instances from three smart-home datasets. Beyond aggregate task
accuracy, we analyze degradation along two dimensions: action components
(operation, device, argument, and value) and task complexity (Simple, Medium,
Complex, Partially Executable, and Infeasible). We find that dense models
exhibit narrow safe pruning regions followed by sharp degradation, whereas
the MoE model tolerates greater expert removal. Pruning also
degrades grounded specificity (i.e., device and value prediction) before
schema-level intent, while baseline task difficulty does not reliably show
pruning sensitivity. Under aggressive dense pruning, models can further shift
toward systematic over-refusal of executable requests.

To sum up, our main contributions are:
(1) a fine-grained framework for analyzing pruning-induced degradation across
action components and smart-home task-complexity levels;
(2) a systematic evaluation across three architecture families, four classes
of structured pruning, and three smart-home datasets with post-pruning SFT;
and (3) empirical characterization of deployment-relevant failure patterns,
including pruning cliffs, specificity-first degradation, task-dependent
pruning sensitivity, and over-refusal under aggressive pruning.

\section{Related Work}
\paragraph{LLM pruning.}
Structured pruning improves LLM inference efficiency by removing redundant structural components, such as channels, Transformer layers, or experts \cite{ma2023llm, gao2024disp}. Existing methods operate at different structural levels: FLAP \cite{an2024fluctuation} and SlimLLM \cite{pmlr-v267-guo25a} prune width-level components such as channels and attention heads, while ShortGPT \cite{men2025shortgpt} and \citet{gromov2025unreasonable} identify and remove redundant Transformer layers. Hybrid approaches such as 2SSP \cite{sandri2025ssp} combine depth and width pruning, and recent methods extend structured pruning to globally optimized settings and MoE models \cite{lasby2026reap, zhang2026mone}. However, most pruning studies evaluate aggregate accuracy, which can obscure capability-specific degradation.


\paragraph{LLMs for smart homes.}
LLMs have increasingly been explored as natural-language interfaces and autonomous agents for smart-home control \cite{chen2026mist}. \citet{rivkin2024aiot} develop an LLM-based agent that reasons over household context and interacts with smart-home devices. HomeBench \cite{li2025homebench} studies valid and invalid instructions involving single and multiple devices. SimuHome \cite{seo2026simuhome} introduces temporal and environment-aware interactions. SMH-Bench \cite{li2026smh} further evaluates environment-grounded reasoning and action across complex smart-home scenarios. These works primarily study unpruned LLMs for smart homes. The reliability of pruned LLMs for smart-home tool execution remains largely unexplored.


\section{Evaluation Framework}
\begin{figure*}
    \centering
    \includegraphics[width=1\linewidth]{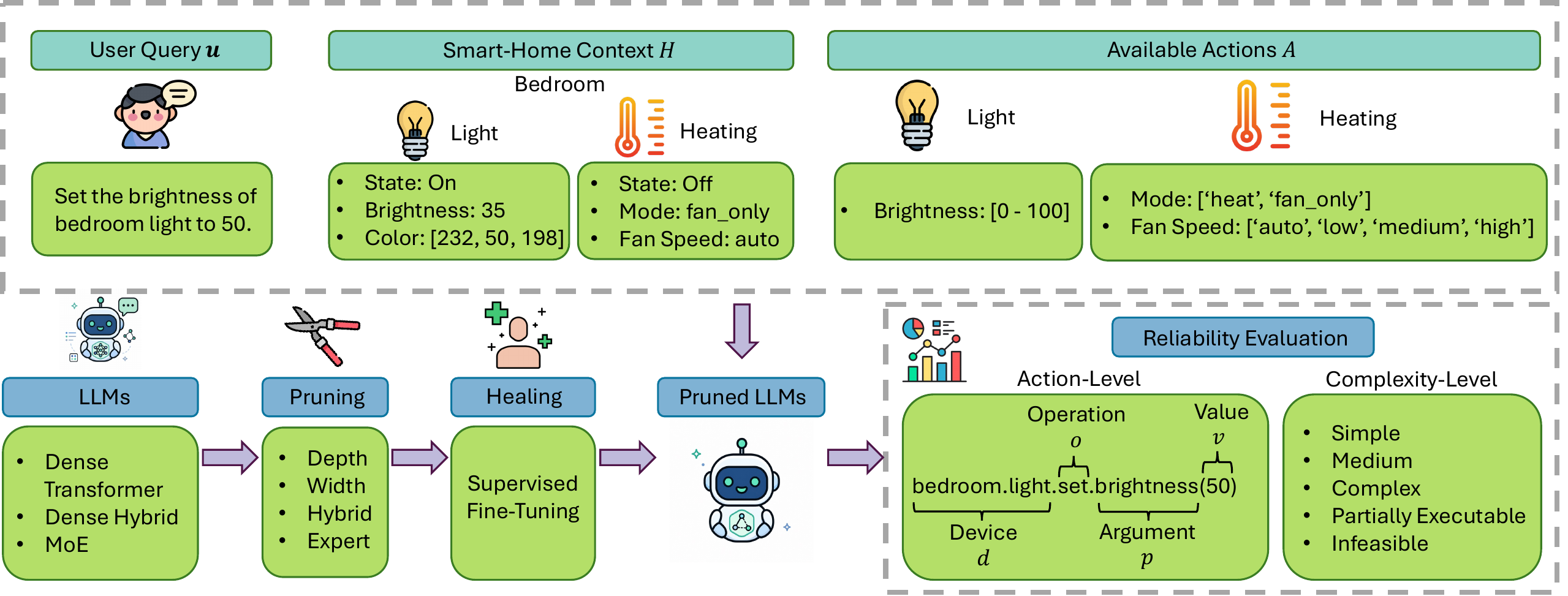}
    \caption{Overview of our framework for evaluating pruning-induced reliability degradation in smart-home tool execution across action components and task-complexity levels.}
    \label{fig_eval_overview}
\end{figure*}

Figure~\ref{fig_eval_overview} summarizes our evaluation framework. We apply
various pruning methods to LLMs with different architectural designs and recover
the pruned models through SFT. We then evaluate reliability
at two complementary granularities. At the action level, we decompose
each generated smart-home operation into its device, operation, argument, and
value components to identify which part of structured execution is most
affected by pruning. At the task level, we stratify requests by
smart-home complexity to characterize when pruning-induced degradation
emerges. Together, these two views allow us to measure not only how much
reliability is lost under pruning, but also what breaks and under which task conditions.

\subsection{Problem Formulation} \label{sec_prob_for}

Following \citet{seo2026simuhome, li2026smh}, we formulate smart-home tool calling as context-grounded structured
prediction. Each instance is represented by $X = \{\boldsymbol{u}, H, A\}$,
where $\boldsymbol{u}$ is the user request, $H$ is the smart-home
context, and $A$ is the set of available device actions.
The context $H$ specifies devices, their capabilities, and current device or environmental states. Given $X$, an LLM produces a set of executions by $
    Y =
    \{
        \boldsymbol{a}_{1}, \boldsymbol{a}_{2}, \ldots, \boldsymbol{a}_{T}
    \}$,
where $T$ is the number of actions $\boldsymbol{a}$ required to fulfill the user request $\boldsymbol{u}$. We define each action $\boldsymbol{a}_j, j= 1,\ldots,T$ into four semantic components $
    \boldsymbol{a}_{j}
    =
    \left(
        o_{j},\;
        d_{j},\;
        p_{j},\;
        v_{j}
    \right)$,
where $o_{j}$ is the operation to be performed, $d_{j}$ is the target-device, $p_{j}$ is the control-argument names, and $v_{j}$ is the argument value. For example, for the request ``dim the bedroom light to 50'', consider the action ``bedroom.light.set.brightness(50)''.
It is represented as $d=$ ``bedroom.light'', $o=$ ``set'', $p=$ ``brightness'', and $v=50$.


\subsection{Smart-Home Task-Complexity Taxonomy}
\label{sec:taxonomy}

To characterize pruning sensitivity across request types, we partition
evaluation instances into five mutually exclusive categories based on
execution complexity and feasibility.

\paragraph{Simple.}
A single-turn request requiring one action on one target device, e.g.,
``turn off the kitchen light.''

\paragraph{Medium.}
A single-turn request requiring multiple actions or multiple target devices,
or an underspecified request that requires clarification but not prior
dialogue context.

\paragraph{Complex.}
A request whose correct execution depends on prior interaction context, such
as resolving anaphoric references or repeating, reversing, or modifying a
previous action.

\paragraph{Partially Executable.}
Only a subset of the requested actions is feasible; the model must execute
the valid portion while rejecting the invalid portion.

\paragraph{Infeasible.}
None of the requested actions is feasible under the current smart-home
context, so the model should avoid execution and reject $X$.

\subsection{LLM Architecture Families}
To examine whether pruning-induced degradation depends on model architecture, we evaluate four models from the Qwen family spanning dense Transformer, dense hybrid, and MoE designs. Qwen3-4B \cite{qwen3technicalreport} is a 4B-parameter dense Transformer with 36 layers and grouped-query attention (GQA). Qwen3.5-4B and Qwen3.5-9B \cite{qwen3.5} adopt a dense hybrid architecture that interleaves Gated DeltaNet linear-attention layers with full-attention layers; both contain 32 layers. Finally, Qwen3.6-35B-A3B \cite{qwen36_35b_a3b} extends this hybrid design with sparse MoE feed-forward layers, comprising 35B total parameters while activating around 3B parameters per token across 40 layers.

This selection enables us to study three complementary factors. Comparing Qwen3-4B with Qwen3.5-4B isolates changes in architectural design at approximately fixed parameter scale; comparing Qwen3.5-4B with Qwen3.5-9B examines the effect of dense model capacity; and Qwen3.6-35B-A3B provides a sparse MoE architecture whose parameter allocation differs fundamentally from dense models. Together, these models allow us to test whether pruning affects smart-home reliability consistently across architectural families or produces architecture-specific failure patterns.









\subsection{Pruning Methods}

We evaluate pruning across four granularities: depth, width,
hybrid, and expert pruning, subject to architectural compatibility.

\paragraph{Depth pruning.}
We use ShortGPT \cite{men2025shortgpt}, which removes low-importance layers
based on Block Influence, and Angular \cite{gromov2025unreasonable}, which
identifies redundant contiguous layer blocks using angular similarity of
boundary representations. Both are applied to Qwen3-4B, Qwen3.5-4B, and
Qwen3.5-9B.

\paragraph{Width pruning.}
We use FLAP \cite{an2024fluctuation}, which prunes feature channels using a
fluctuation-based importance criterion with adaptive sparsity allocation and
bias compensation. We apply FLAP to Qwen3-4B; its method design
does not directly support the Gated DeltaNet modules used in Qwen3.5.

\paragraph{Hybrid pruning.}
We use 2SSP \cite{sandri2025ssp}, which combines feed-forward neuron pruning
with attention-submodule pruning. We apply 2SSP to Qwen3-4B; its
attention-pruning stage assumes conventional attention modules and would
require architecture-specific modifications for Qwen3.5.

\paragraph{Expert pruning.}
For Qwen3.6-35B-A3B, we use REAP \cite{lasby2026reap}, which ranks experts
using router scores and activation magnitudes and removes those with low
estimated contribution.

After pruning, all models undergo the same SFT procedure while keeping the
pruned architecture fixed. Overall, Qwen3-4B is evaluated with ShortGPT,
Angular, FLAP, and 2SSP; Qwen3.5-4B and Qwen3.5-9B with ShortGPT and Angular;
and Qwen3.6-35B-A3B with REAP.

\subsection{Pruning-Degradation Evaluation}
\label{sec:metrics}
Let $M_0$ denote the original SFT LLM without pruning and $M_{\pi}$ an LLM
obtained under pruning configuration $\pi$, which specifies the pruning
method and ratio. We evaluate performance at
two complementary granularities: action-component level, which
identifies what parts of structured tool execution are affected by pruning,
and task-complexity level, which identifies under what task
conditions degradation occurs. For evaluation instance $X_i, i=1,\ldots,N$, let $Y_i$ denote the ground-truth action
set and $M(X_i)$ the action set produced by model $M$. Overall task
accuracy is $\mathrm{Acc}(M) =\frac{1}{N}\sum_{i=1}^{N} \mathbb{I}\!\left[M(X_i)=Y_i\right]$. An instance is considered correct only when the complete required execution of $a$ is satisfied. For multi-action requests, actions are
matched independently of their order.
\paragraph{Action-component accuracy.}
We evaluate the four
action components $\{o,d,p,v\}$ defined in Section~\ref{sec_prob_for}. For instance $X_i$, let $Y_i^{(j)}$ and $M^{(j)}(X_i)$ denote its ground-truth and
predicted component from model $M$, where $j \in \{o,d,p,v\}$, respectively. Component $j$'s accuracy of $M$ is $\mathrm{Acc}^{(j)}(M)
    =
    \frac{1}{N}
    \sum_{i=1}^{N}
    \mathbb{I}\!\left[
        M^{(j)}(X_i)=Y_i^{(j)}
    \right]$.
    
\paragraph{Complexity-level accuracy.}
We report task accuracy
separately for each category $c \in \{$Simple, Medium, Complex, Partially Executable, Infeasible$\}$ in the complexity taxonomy of
Section~\ref{sec:taxonomy}. Let $I_c$ denote the set of
evaluation instances belonging to category $c$. We define the category $c$'s accuracy of $M$ as $\mathrm{Acc}_{c}(M)
    =
    \frac{1}{|I_c|}
    \sum_{i\in I_c}
    \mathbb{I}\!\left[M(X_i)=Y_i\right]$.
    

\paragraph{Pruning-induced degradation.}
We measure reliability degradation relative to the corresponding
unpruned model: $\Delta\mathrm{Acc}(M_{\pi}) =
    \mathrm{Acc}(M_{\pi})-\mathrm{Acc}(M_0), 
    \Delta\mathrm{Acc}^{(k)}(M_{\pi}) =
    \mathrm{Acc}^{(k)}(M_{\pi})
    -
    \mathrm{Acc}^{(k)}(M_0), 
    \Delta\mathrm{Acc}_{c}(M_{\pi}) =
    \mathrm{Acc}_{c}(M_{\pi})
    -
    \mathrm{Acc}_{c}(M_0)$.
We report these changes with negative values
indicating degradation after pruning. Action-component changes
identify which parts of structured tool execution are most sensitive
to capacity reduction, while complexity-level changes identify the
task conditions under which these failures are most pronounced.

\section{Experiments}

\subsection{Experimental Setup}
\label{sec:datasets}
For SFT and evaluation datasets, we use the training and test splits, respectively,
from the same three sources: proprietary smart-home tool-calling
dialogues (SHTC), HomeBench \cite{li2025homebench}, and
Home-Assistant-Requests-V2 (HAR)\footnote{\url{https://huggingface.co/datasets/acon96/Home-Assistant-Requests-V2}}.
All pruned models are healed using the same SFT mixture constructed from
the training splits. Table~\ref{tab:cat_counts} shows
the statistics of the evaluation set. For SHTC, to comply
with proprietary-data requirements, we report only performance
changes $\Delta\mathrm{Acc}$ relative to the corresponding unpruned model $M_0$. Additional details of experimental setup are provided in Appendix \ref{appendix_exp_set}.

\begin{table}
\centering
\small
\setlength{\tabcolsep}{4pt}
\resizebox{\columnwidth}{!}{%
\begin{tabular}{lrrr}
\toprule
\textbf{Complexity Category} & \textbf{SHTC (\%)} & \textbf{HomeBench} & \textbf{HAR} \\
\midrule
Simple               & 66.6\% & 6{,}187 & 1{,}903 \\
Medium               & 29.4\% &   245   &   291   \\
Complex              &  3.9\% &    --   &    --   \\
Partially Executable &    --  & 4{,}072 &    --   \\
Infeasible           &    --  & 6{,}862 &    --   \\
\midrule
Total                & 100\%  & 17{,}366 & 2{,}194 \\
\bottomrule
\end{tabular}%
}
\caption{Statistics of the evaluation sets. For SHTC, we report the proportion of instances in each category.}
\label{tab:cat_counts}
\end{table}

\subsection{Results}
\begin{figure*}
\centering
\begin{subfigure}[t]{0.24\textwidth}\centering
\includegraphics[width=\linewidth]{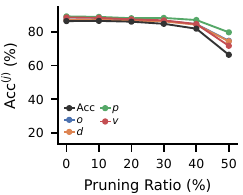}
\caption{ShortGPT}
\end{subfigure}
\hfill
\begin{subfigure}[t]{0.24\textwidth}\centering
\includegraphics[width=\linewidth]{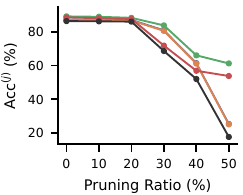}
\caption{Angular}
\end{subfigure}
\hfill
\begin{subfigure}[t]{0.24\textwidth}\centering
\includegraphics[width=\linewidth]{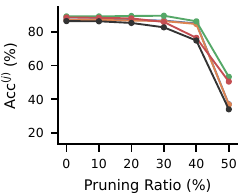}
\caption{FLAP}
\end{subfigure}
\hfill
\begin{subfigure}[t]{0.24\textwidth}\centering
\includegraphics[width=\linewidth]{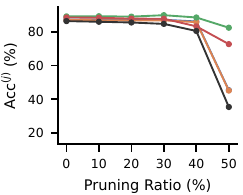}
\caption{2SSP}
\end{subfigure}
\caption{HomeBench action-component accuracy across pruning ratios for four pruning methods in Qwen3-4B.}
\label{fig:main_facets_homebench_qwen3_4b}
\end{figure*}

We show HomeBench action-component and complexity-level accuracy for
Qwen3-4B in
Figures~\ref{fig:main_facets_homebench_qwen3_4b}
and~\ref{fig:main_cats_homebench_qwen3_4b}. Complete results are provided in Appendix
Figures~\ref{fig:panels_facets_homebench}-\ref{fig:panels_facets_greenv5}
and~\ref{fig:panels_cats_homebench}-\ref{fig:panels_cats_greenv5}. Here, we focus on the key findings and insights.

\begin{figure*}
\centering
\begin{subfigure}[t]{0.24\textwidth}\centering
\includegraphics[width=\linewidth]{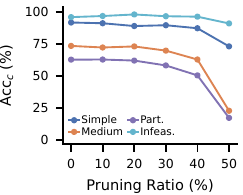}
\caption{ShortGPT}
\end{subfigure}
\hfill
\begin{subfigure}[t]{0.24\textwidth}\centering
\includegraphics[width=\linewidth]{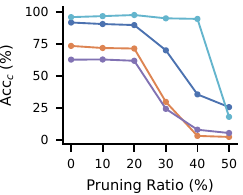}
\caption{Angular}
\end{subfigure}
\hfill
\begin{subfigure}[t]{0.24\textwidth}\centering
\includegraphics[width=\linewidth]{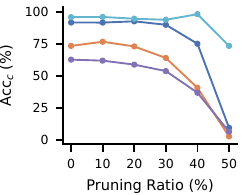}
\caption{FLAP}
\end{subfigure}
\hfill
\begin{subfigure}[t]{0.24\textwidth}\centering
\includegraphics[width=\linewidth]{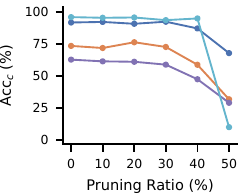}
\caption{2SSP}
\end{subfigure}
\caption{HomeBench complexity-level accuracy across pruning ratios for four pruning methods in Qwen3-4B.}
\label{fig:main_cats_homebench_qwen3_4b}
\end{figure*}



\begin{figure*}[t]
    \centering

    \begin{minipage}[t]{0.32\textwidth}
        \centering
        \includegraphics[width=0.9\linewidth]{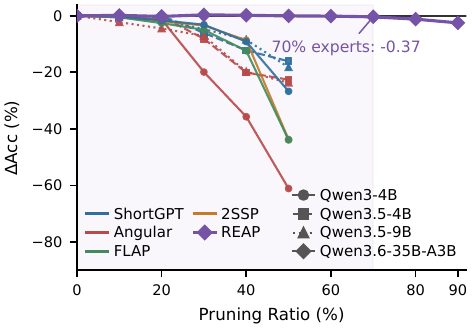}
        \captionof{figure}{Average $\Delta \mathrm{Acc}$ of different LLMs
        under different pruning methods across three datasets.}
        \label{fig:pruning-regimes}
    \end{minipage}
    \hfill
    \begin{minipage}[t]{0.32\textwidth}
        \centering
        \includegraphics[width=0.9\linewidth]{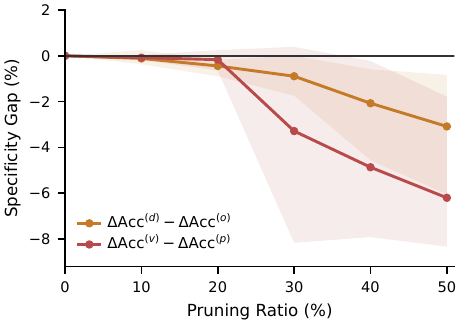}
        \captionof{figure}{Specificity gaps
        $\Delta\mathrm{Acc}^{(d)}-\Delta\mathrm{Acc}^{(o)}$ and
        $\Delta\mathrm{Acc}^{(v)}-\Delta\mathrm{Acc}^{(p)}$ across dense
        pruning configurations.}
        \label{fig:specificity-gap-main}
    \end{minipage}
    \hfill
    \begin{minipage}[t]{0.32\textwidth}
        \centering
        \includegraphics[width=0.9\linewidth]{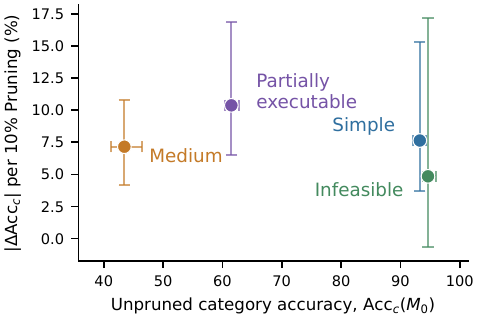}
        \captionof{figure}{$|\Delta \mathrm{Acc}_c|$ per additional 10\%
        pruning versus mean $\mathrm{Acc}_c(M_0)$ across public datasets.}
        \label{fig:category-fragility-main}
    \end{minipage}

\end{figure*}

\paragraph{Dense pruning exhibits a narrower safe region than MoE pruning.}
Figure~\ref{fig:pruning-regimes} contrasts the dense and MoE pruning
curves. Overall, dense pruning exhibits sharp cliffs, whereas MoE pruning
maintains a wider plateau. At 10\% dense pruning, seven of eight LLM-method
combinations lose less than 0.6\% accuracy, suggesting that mild pruning mainly
removes redundant capacity. 
Pruning ratio alone does not determine retained capability; which structures
are removed is also critical. The dense hybrid Qwen3.5-4B is more robust than
the same-scale GQA model, possibly because its heterogeneous Gated DeltaNet and
full-attention layers provide more alternative information pathways. However,
increasing the hybrid model from 4B to 9B yields no consistent benefit,
suggesting that additional parameters do not necessarily correspond to
pruning-redundant capacity. For REAP, removing 70\% of experts reduces stored parameters from 35B to
approximately 12.5B with a nonsignificant $-0.37\%$ change. Unlike dense pruning, REAP
reduces stored experts while preserving routing and eight active experts per
token (around 3B parameters), which explains its wider safe region.


\paragraph{Pruning removes specificity before intent.}
We distinguish schema-level intent, represented by operation and argument
$(o,p)$, from grounded specificity, represented by device and value $(d,v)$.
We compare device with operation because $o$ specifies what to do and $d$
where to do it; similarly, $p$ identifies the controlled attribute and $v$
its requested setting. Figure~\ref{fig:specificity-gap-main} plots
$\Delta\mathrm{Acc}^{(d)}-\Delta\mathrm{Acc}^{(o)}$ and
$\Delta\mathrm{Acc}^{(v)}-\Delta\mathrm{Acc}^{(p)}$, averaged across dense
pruning results over all three datasets; negative values indicate greater
degradation of specificity than intent. Both gaps become increasingly negative
with pruning. At 50\%, mean losses are 24.2\% vs.\ 21.1\% for $d$ vs.\ $o$,
and 20.7\% vs.\ 14.5\% for $v$ vs.\ $p$, with both specificity components
degrading at least as much as their paired intent components. It is because $o$ and $p$
come from a relatively small, repeated schema, whereas $d$ and $v$
require instance-specific grounding in the provided context and precise
selection among candidates. $d$ is also the weakest unpruned component, suggesting that pruning amplifies an already difficult
grounding step. Thus, reduced capacity appears to preserve the general action
schema longer than the contextual details needed to execute it precisely.

\begin{wrapfigure}{r}{0.65\columnwidth}
    \centering
    \includegraphics[width=\linewidth]{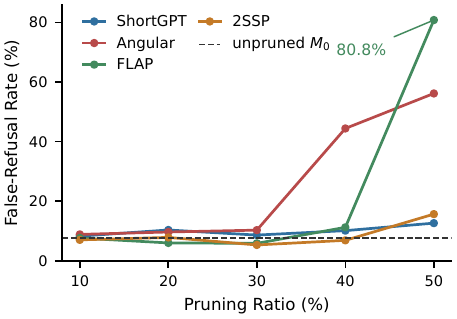}
    \caption{False-refusal rate on executable
    HomeBench requests.}
    \label{fig:false-refusal-main}
\end{wrapfigure}

\paragraph{Baseline accuracy does not predict pruning sensitivity.}
Figure~\ref{fig:category-fragility-main} shows that pruning sensitivity is
not monotonic in unpruned accuracy. Partially Executable requests are
the most sensitive, losing about 10.3\% accuracy per additional 10\% pruning, followed by Simple (7.7\%) and Medium (7.2\%) requests. Notably,
Simple and Infeasible requests have similarly high unpruned accuracy, yet Simple requests degrade faster than
Infeasible requests. Conversely, Medium requests begin at much lower
accuracy but exhibit pruning sensitivity comparable to Simple
requests. These contrasts show that baseline difficulty alone does not
determine robustness to capacity reduction. Partially Executable requests
are especially vulnerable because they require jointly identifying feasible
and infeasible portions while coordinating execution and rejection. Simple
requests, despite being easy for the unpruned model, still require precise
grounding to a device and action, which can deteriorate under pruning.
Infeasible requests instead primarily require withholding action; together
with Figure~\ref{fig:false-refusal-main}, their relative robustness may
partly reflect pruning-induced preference for rejection.

\paragraph{Deep pruning can collapse into over-refusal.}
Figure~\ref{fig:false-refusal-main} shows that aggressive pruning increasingly
shifts the model from incorrect execution toward refusing to act using Qwen3-4B. On executable
HomeBench requests, unpruned Qwen3-4B refuses 7.7\%; at 50\% pruning this
rises to 56.2\% with Angular and 80.8\% with FLAP. Yet these models still
withhold action on 86.8\% and 99.9\% of infeasible requests, respectively.
One possible explanation is that generating a valid tool call requires precise
device, operation, and value grounding, whereas refusal is a simpler
low-commitment output. As pruning weakens grounded generation, the model may
therefore increasingly default to rejection when uncertain. Consequently, an
infeasible-only refusal metric would make the FLAP model appear robust while it
rejects four out of five valid commands. Deep pruning thus changes the model's
decision policy, not merely its execution accuracy.



\section{Discussion and Future Work}
\label{sec:discussion}

Our results show that pruning in smart-home tool calling must preserve more than
general semantic capability. Unlike conventional language or structured-prediction
tasks, correct execution is conditioned on the current complex home-environment. Operation and argument
prediction remain stable longer than device and value prediction, indicating
that pruning can preserve the general action schema while damaging the contextual
specificity. Partially Executable requests are especially
sensitive, and aggressive pruning can shift models toward
over-refusal. Thus, aggregate accuracy alone is insufficient in smart homes.
For bounded-domain MoEs, expert pruning is a promising starting point. For dense LLMs, pruning should be more conservative, with multiple ratios and criteria evaluated for degradation.



In smart homes, future work should develop pruning and recovery methods that explicitly preserve
environment grounding. Pruning criteria could protect components important for
device resolution and precise value prediction rather than
relying only on generic importance scores. Healing data should target
pruning-specific failures, including distractor-rich device resolution, precise
values, multi-device requests, and partially executable instructions. Because
aggressive pruning can induce over-refusal, recovery should also calibrate the
execute-versus-reject decision using feasible and infeasible requests. Unlike
conventional domains where recovery may mainly restore aggregate language or
reasoning performance, smart-home healing must restore the link between language,
the current environment, and selective action.

\section{Conclusion}
We systematically study how pruning affects LLMs for context-grounded smart-home tool calling across architectures, pruning methods, and severities. Beyond aggregate accuracy, we examine degradation at the action-component and task-complexity levels. We find that grounded specificity degrades before schema-level intent, and aggressive dense pruning can induce over-refusal. These findings show that pruning should be evaluated not only by overall accuracy, but also by architecture, workload, and failure mode. Fine-grained evaluation is therefore important for identifying pruning regimes that reduce model cost while preserving reliable smart-home tool execution.

\section*{Limitations}
Despite the insights, our study has several limitations. First, although we cover dense
Transformer, dense hybrid, and MoE architectures, all evaluated models
belong to the Qwen family; whether the observed pruning patterns generalize
to other model families remains to be studied. Second, pruning methods are
not evaluated in a fully factorial manner because some methods are
architecture-specific; for example, FLAP and 2SSP are not directly
applicable to the Gated DeltaNet-based models. Finally, our evaluation is restricted to smart-home tool calling, and some
complexity categories are available in only a subset of the three datasets,
which limits broader generalization across domains. Future work could extend the analysis to
additional model families, domains, pruning methods, and deployment-level
efficiency measures such as latency and throughput.




\bibliography{custom}

\newpage

\appendix

\section{Additional Experimental Setup}\label{appendix_exp_set}
\paragraph{Pruning Ratios.} 
Dense LLM architectures are pruned at ratios from $10\%$ to $50\%$ in increments of $10\%$. For the MoE architecture, we evaluate more aggressive expert pruning, removing $10\%$ to $90\%$ of experts to account for its higher structural sparsity.

\paragraph{SFT Data.} The SFT data mixture is balanced three ways by source rather than pooled in natural
proportion: $16{,}667$ examples each from SHTC, HomeBench and
HAR, giving $50{,}001$ training examples, with $100$ further examples per source
held out for validation.

\paragraph{Training Settings.} SFT is performed for one epoch with a global batch size of $32$ and a micro-batch size of $1$, resulting in $1{,}563$ optimizer steps. We use a maximum sequence length of $32{,}768$ tokens and bf16 mixed precision. Optimization uses Adam with $\beta_2{=}0.98$ and an initial learning rate of $5\times10^{-6}$, with $50$ warmup steps followed by decay to a minimum learning rate of $5\times10^{-7}$. We use $32$ NVIDIA A100 80 GB GPUs for dense LLM architectures and $64$ NVIDIA A100 80 GB GPUs for the MoE model.

\paragraph{Generation Settings.} 
Unless stated otherwise, we use greedy decoding with temperature as $0$. To quantify the extent to which per-instance variation may arise from sampling stochasticity, we additionally evaluate Qwen-recommended sampling settings with temperature as $0.7$, $\mathrm{top\text{-}}p=0.8$, and $\mathrm{top\text{-}}k=20$.






\section{Prompt Example}
\label{sec:appendix_prompt}
\begin{figure*}
    \centering
\includegraphics[width=1\linewidth]{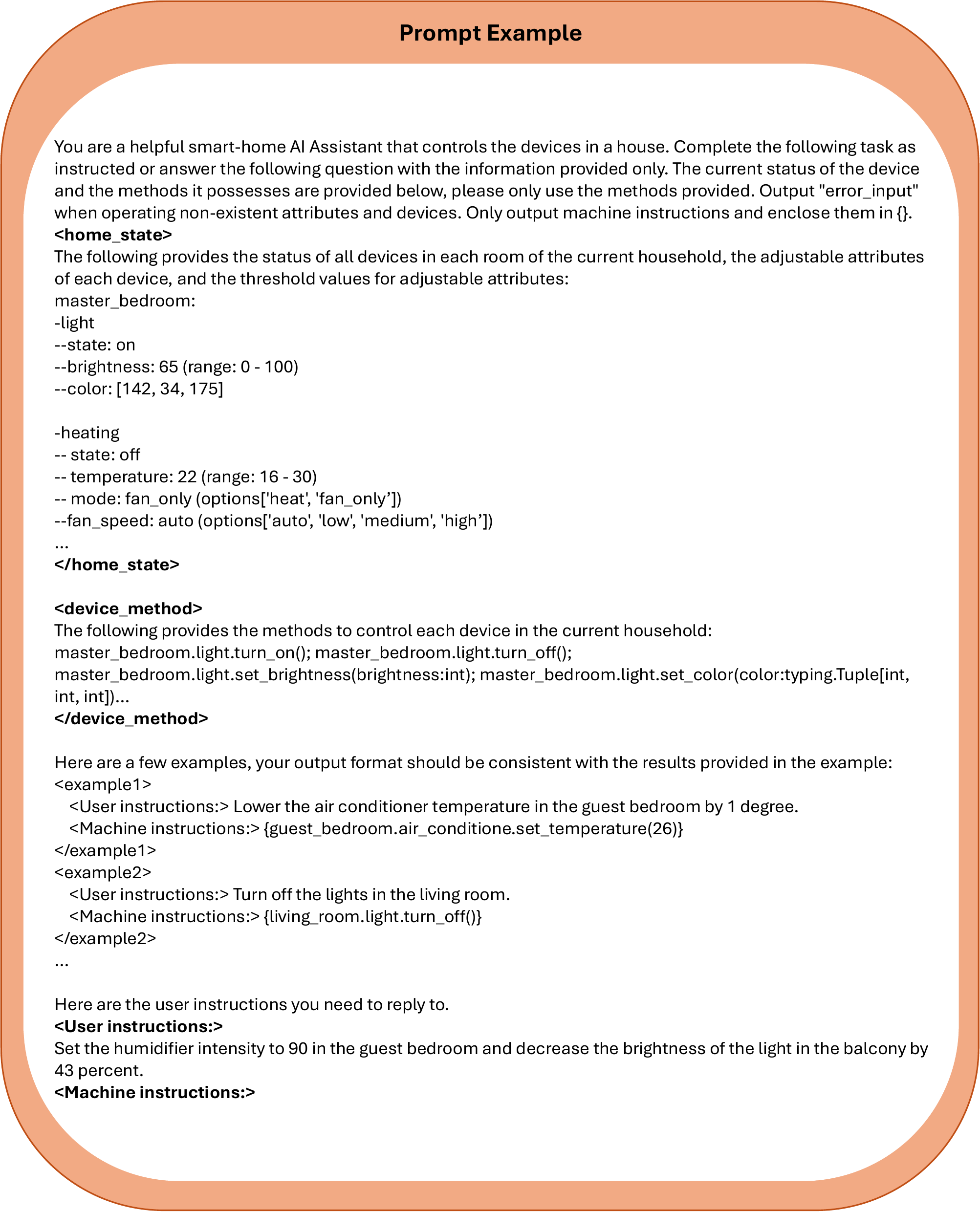}
    \caption{Prompt example for HomeBench.}
    \label{fig_app_prompt}
\end{figure*}

Figure~\ref{fig_app_prompt} shows an example prompt used for smart-home
tool calling in HomeBench. The prompt provides the LLM with the current smart-home
state, including available devices, their attributes, and valid value ranges,
together with the device-control methods that may be invoked. The prompt also
specifies the required machine-instruction output format. The target user request is
then appended at the end of the prompt, and the model is required to generate
the corresponding executable action(s) using only the provided devices and
methods. For unsupported devices or attributes, the prompt instructs the
model to return ``error\_input''.

\section{Action-Component Accuracy}
\label{app:action-component}
\begin{figure*}
\centering
\begin{subfigure}[t]{0.315\textwidth}\centering
\includegraphics[width=\linewidth]{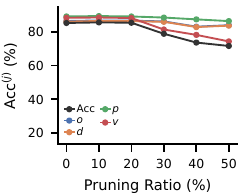}
\caption{Qwen3.5-4B, ShortGPT}
\end{subfigure}
\hfill
\begin{subfigure}[t]{0.315\textwidth}\centering
\includegraphics[width=\linewidth]{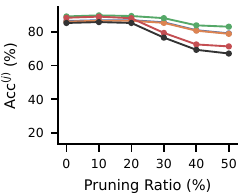}
\caption{Qwen3.5-4B, Angular}
\end{subfigure}
\hfill
\begin{subfigure}[t]{0.315\textwidth}\centering
\includegraphics[width=\linewidth]{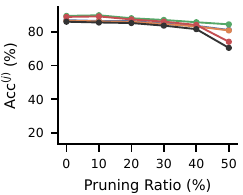}
\caption{Qwen3.5-9B, ShortGPT}
\end{subfigure}
\\
\begin{subfigure}[t]{0.315\textwidth}\centering
\includegraphics[width=\linewidth]{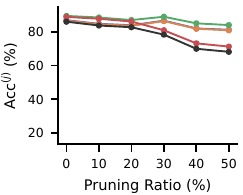}
\caption{Qwen3.5-9B, Angular}
\end{subfigure}
\begin{subfigure}[t]{0.315\textwidth}\centering
\includegraphics[width=\linewidth]{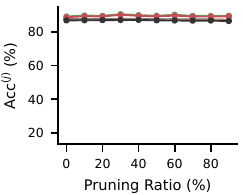}
\caption{Qwen3.6-35B-A3B, REAP}
\end{subfigure}
\caption{\textbf{HomeBench: action-component accuracy for the remaining
LLM-pruning combinations.} $\mathrm{Acc}$ together with the four action components $o,d,p,v$ against pruning ratio. Each subfigure is one LLM and pruning method. 
Qwen3-4B results are in
Figure~\ref{fig:main_facets_homebench_qwen3_4b}.}
\label{fig:panels_facets_homebench}
\end{figure*}

\begin{figure*}
\centering
\begin{subfigure}[t]{0.315\textwidth}\centering
\includegraphics[width=\linewidth]{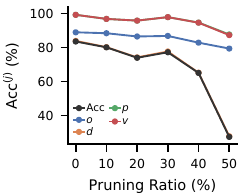}
\caption{Qwen3-4B, ShortGPT}
\end{subfigure}
\hfill
\begin{subfigure}[t]{0.315\textwidth}\centering
\includegraphics[width=\linewidth]{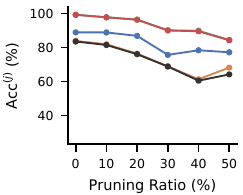}
\caption{Qwen3-4B, Angular}
\end{subfigure}
\hfill
\begin{subfigure}[t]{0.315\textwidth}\centering
\includegraphics[width=\linewidth]{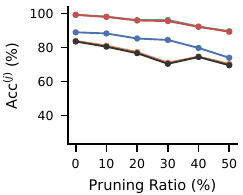}
\caption{Qwen3-4B, FLAP}
\end{subfigure}
\\[2pt]
\begin{subfigure}[t]{0.315\textwidth}\centering
\includegraphics[width=\linewidth]{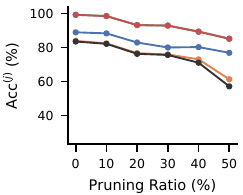}
\caption{Qwen3-4B, 2SSP}
\end{subfigure}
\hfill
\begin{subfigure}[t]{0.315\textwidth}\centering
\includegraphics[width=\linewidth]{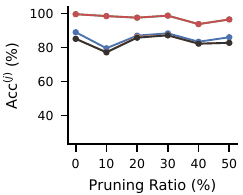}
\caption{Qwen3.5-4B, ShortGPT}
\end{subfigure}
\hfill
\begin{subfigure}[t]{0.315\textwidth}\centering
\includegraphics[width=\linewidth]{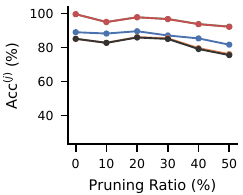}
\caption{Qwen3.5-4B, Angular}
\end{subfigure}
\\[2pt]
\begin{subfigure}[t]{0.315\textwidth}\centering
\includegraphics[width=\linewidth]{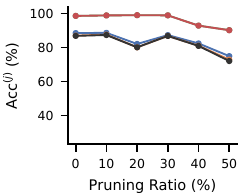}
\caption{Qwen3.5-9B, ShortGPT}
\end{subfigure}
\hfill
\begin{subfigure}[t]{0.315\textwidth}\centering
\includegraphics[width=\linewidth]{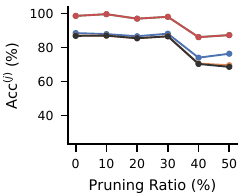}
\caption{Qwen3.5-9B, Angular}
\end{subfigure}
\hfill
\begin{subfigure}[t]{0.315\textwidth}\centering
\includegraphics[width=\linewidth]{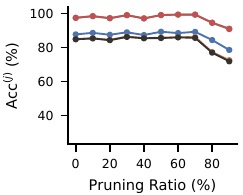}
\caption{Qwen3.6-35B-A3B, REAP}
\end{subfigure}
\caption{\textbf{HAR: action-component accuracy.} $\mathrm{Acc}$ together with the four action components $o,d,p,v$ against pruning ratio.
Each subfigure is one LLM and pruning method.}
\label{fig:panels_facets_har}
\end{figure*}

\begin{figure*}[t]
\centering
\begin{subfigure}[t]{0.315\textwidth}\centering
\includegraphics[width=\linewidth]{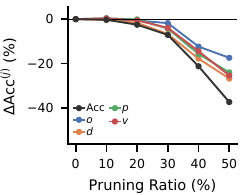}
\caption{Qwen3-4B, ShortGPT}
\end{subfigure}
\hfill
\begin{subfigure}[t]{0.315\textwidth}\centering
\includegraphics[width=\linewidth]{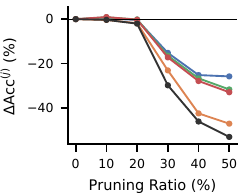}
\caption{Qwen3-4B, Angular}
\end{subfigure}
\hfill
\begin{subfigure}[t]{0.315\textwidth}\centering
\includegraphics[width=\linewidth]{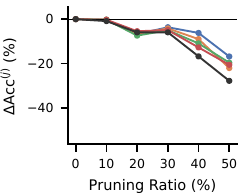}
\caption{Qwen3-4B, FLAP}
\end{subfigure}
\\[2pt]
\begin{subfigure}[t]{0.315\textwidth}\centering
\includegraphics[width=\linewidth]{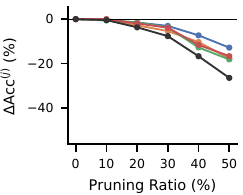}
\caption{Qwen3-4B, 2SSP}
\end{subfigure}
\hfill
\begin{subfigure}[t]{0.315\textwidth}\centering
\includegraphics[width=\linewidth]{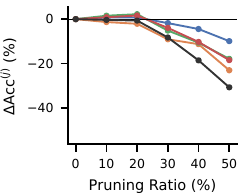}
\caption{Qwen3.5-4B, ShortGPT}
\end{subfigure}
\hfill
\begin{subfigure}[t]{0.315\textwidth}\centering
\includegraphics[width=\linewidth]{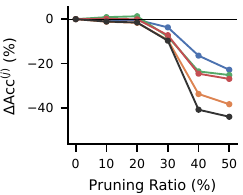}
\caption{Qwen3.5-4B, Angular}
\end{subfigure}
\\[2pt]
\begin{subfigure}[t]{0.315\textwidth}\centering
\includegraphics[width=\linewidth]{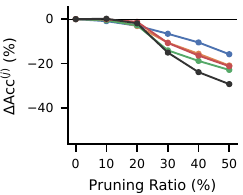}
\caption{Qwen3.5-9B, ShortGPT}
\end{subfigure}
\hfill
\begin{subfigure}[t]{0.315\textwidth}\centering
\includegraphics[width=\linewidth]{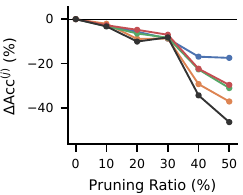}
\caption{Qwen3.5-9B, Angular}
\end{subfigure}
\hfill
\begin{subfigure}[t]{0.315\textwidth}\centering
\includegraphics[width=\linewidth]{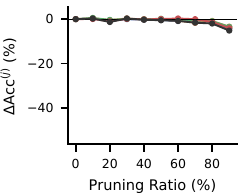}
\caption{Qwen3.6-35B-A3B, REAP}
\end{subfigure}
\caption{\textbf{SHTC: action-component accuracy.} $\Delta\mathrm{Acc}$ relative to $M_0$ together with the four action components $o,d,p,v$ against pruning ratio.
Each subfigure is one LLM and pruning method.}
\label{fig:panels_facets_greenv5}
\end{figure*}

Figures~\ref{fig:panels_facets_homebench}-\ref{fig:panels_facets_greenv5} provide the complete action-component results across datasets,
LLMs, pruning methods, and pruning ratios. Overall, the results
support the pattern discussed in Section~4.2: dense models generally preserve
all four components under mild pruning, followed by increasingly
component-specific degradation at higher pruning ratios. In particular,
device ($d$) and value ($v$) accuracy often decline more rapidly than
operation ($o$) and argument ($p$) accuracy, indicating that pruning tends to
damage grounded specificity before schema-level intent. This behavior is consistent across HomeBench (Figure~\ref{fig:panels_facets_homebench}), HAR (Figure~\ref{fig:panels_facets_har}), and
SHTC (Figure~\ref{fig:panels_facets_greenv5}). In contrast, the MoE
model pruned with REAP maintains relatively stable component-level performance
over a much wider range of pruning ratios, with noticeable degradation
appearing primarily under the most aggressive expert pruning.

\section{Complexity-Level Accuracy}
\label{app:complexity-level}
\begin{figure*}[t]
\centering
\begin{subfigure}[t]{0.315\textwidth}\centering
\includegraphics[width=\linewidth]{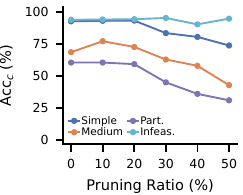}
\caption{Qwen3.5-4B, ShortGPT}
\end{subfigure}
\hfill
\begin{subfigure}[t]{0.315\textwidth}\centering
\includegraphics[width=\linewidth]{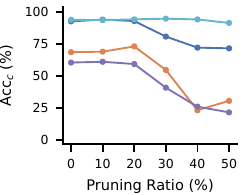}
\caption{Qwen3.5-4B, Angular}
\end{subfigure}
\hfill
\begin{subfigure}[t]{0.315\textwidth}\centering
\includegraphics[width=\linewidth]{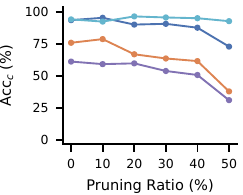}
\caption{Qwen3.5-9B, ShortGPT}
\end{subfigure}
\\
\begin{subfigure}[t]{0.315\textwidth}\centering
\includegraphics[width=\linewidth]{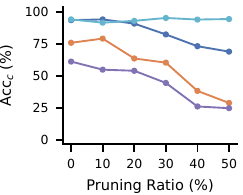}
\caption{Qwen3.5-9B, Angular}
\end{subfigure}
\begin{subfigure}[t]{0.315\textwidth}\centering
\includegraphics[width=\linewidth]{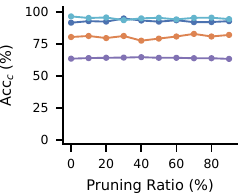}
\caption{Qwen3.6-35B-A3B, REAP}
\end{subfigure}
\caption{\textbf{HomeBench: accuracy of task-complexity categories for the remaining LLM-pruning combinations.} $\mathrm{Acc}_c$ for the complexity categories present in the source against pruning ratio. Each subfigure is one LLM and pruning method. Qwen3-4B results are in
Figure~\ref{fig:main_cats_homebench_qwen3_4b}.}
\label{fig:panels_cats_homebench}
\end{figure*}

\begin{figure*}[t]
\centering
\begin{subfigure}[t]{0.315\textwidth}\centering
\includegraphics[width=\linewidth]{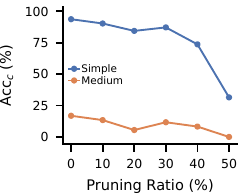}
\caption{Qwen3-4B, ShortGPT}
\end{subfigure}
\hfill
\begin{subfigure}[t]{0.315\textwidth}\centering
\includegraphics[width=\linewidth]{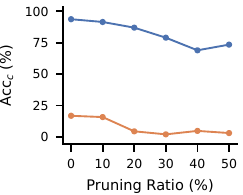}
\caption{Qwen3-4B, Angular}
\end{subfigure}
\hfill
\begin{subfigure}[t]{0.315\textwidth}\centering
\includegraphics[width=\linewidth]{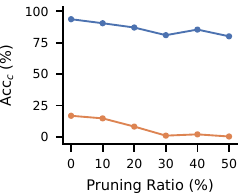}
\caption{Qwen3-4B, FLAP}
\end{subfigure}
\\[2pt]
\begin{subfigure}[t]{0.315\textwidth}\centering
\includegraphics[width=\linewidth]{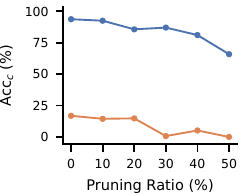}
\caption{Qwen3-4B, 2SSP}
\end{subfigure}
\hfill
\begin{subfigure}[t]{0.315\textwidth}\centering
\includegraphics[width=\linewidth]{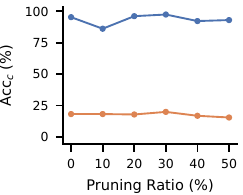}
\caption{Qwen3.5-4B, ShortGPT}
\end{subfigure}
\hfill
\begin{subfigure}[t]{0.315\textwidth}\centering
\includegraphics[width=\linewidth]{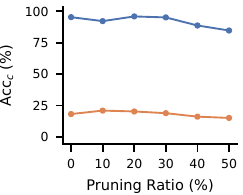}
\caption{Qwen3.5-4B, Angular}
\end{subfigure}
\\[2pt]
\begin{subfigure}[t]{0.315\textwidth}\centering
\includegraphics[width=\linewidth]{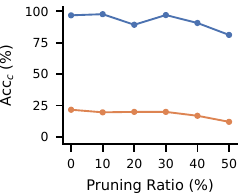}
\caption{Qwen3.5-9B, ShortGPT}
\end{subfigure}
\hfill
\begin{subfigure}[t]{0.315\textwidth}\centering
\includegraphics[width=\linewidth]{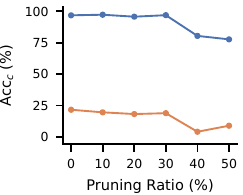}
\caption{Qwen3.5-9B, Angular}
\end{subfigure}
\hfill
\begin{subfigure}[t]{0.315\textwidth}\centering
\includegraphics[width=\linewidth]{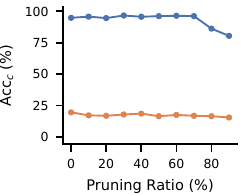}
\caption{Qwen3.6-35B-A3B, REAP}
\end{subfigure}
\caption{\textbf{HAR: accuracy of task-complexity categories.} $\mathrm{Acc}_c$ for the complexity categories present in the source against pruning ratio. Each subfigure is one LLM and pruning method.}
\label{fig:panels_cats_har}
\end{figure*}

\begin{figure*}[t]
\centering
\begin{subfigure}[t]{0.315\textwidth}\centering
\includegraphics[width=\linewidth]{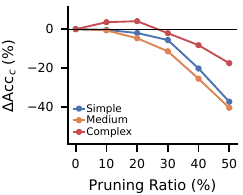}
\caption{Qwen3-4B, ShortGPT}
\end{subfigure}
\hfill
\begin{subfigure}[t]{0.315\textwidth}\centering
\includegraphics[width=\linewidth]{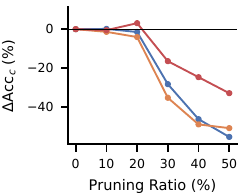}
\caption{Qwen3-4B, Angular}
\end{subfigure}
\hfill
\begin{subfigure}[t]{0.315\textwidth}\centering
\includegraphics[width=\linewidth]{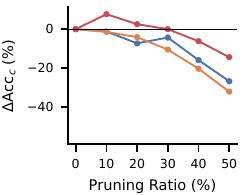}
\caption{Qwen3-4B, FLAP}
\end{subfigure}
\\[2pt]
\begin{subfigure}[t]{0.315\textwidth}\centering
\includegraphics[width=\linewidth]{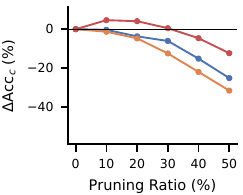}
\caption{Qwen3-4B, 2SSP}
\end{subfigure}
\hfill
\begin{subfigure}[t]{0.315\textwidth}\centering
\includegraphics[width=\linewidth]{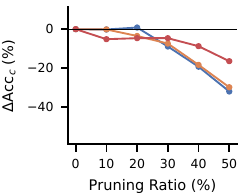}
\caption{Qwen3.5-4B, ShortGPT}
\end{subfigure}
\hfill
\begin{subfigure}[t]{0.315\textwidth}\centering
\includegraphics[width=\linewidth]{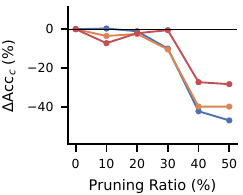}
\caption{Qwen3.5-4B, Angular}
\end{subfigure}
\\[2pt]
\begin{subfigure}[t]{0.315\textwidth}\centering
\includegraphics[width=\linewidth]{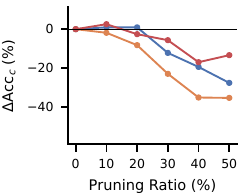}
\caption{Qwen3.5-9B, ShortGPT}
\end{subfigure}
\hfill
\begin{subfigure}[t]{0.315\textwidth}\centering
\includegraphics[width=\linewidth]{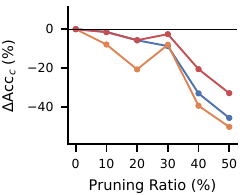}
\caption{Qwen3.5-9B, Angular}
\end{subfigure}
\hfill
\begin{subfigure}[t]{0.315\textwidth}\centering
\includegraphics[width=\linewidth]{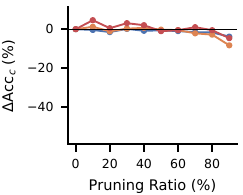}
\caption{Qwen3.6-35B-A3B, REAP}
\end{subfigure}
\caption{\textbf{SHTC: accuracy of task-complexity categories.} $\Delta\mathrm{Acc}$ relative to $M_0$ for the complexity categories present in the source against pruning ratio. Each subfigure is one LLM and pruning method.}
\label{fig:panels_cats_greenv5}
\end{figure*}

Figures~\ref{fig:panels_cats_homebench}-\ref{fig:panels_cats_greenv5} present the complete complexity-level results across datasets,
LLM architectures, pruning methods, and pruning ratios. The results further
show that pruning sensitivity is not determined
by the unpruned accuracy alone. For HomeBench (Figure~\ref{fig:panels_cats_homebench}), Medium and Partially Executable
requests generally degrade more rapidly under aggressive dense pruning, while
Infeasible requests remain comparatively robust. The HAR results (Figure~\ref{fig:panels_cats_har}) similarly
show substantial degradation of Medium requests for several dense-pruning
configurations, although the magnitude varies across architectures and methods. For SHTC (Figure~\ref{fig:panels_cats_greenv5}), Simple and Medium requests generally experience larger losses than
Complex requests at high dense-pruning ratios, further illustrating that
baseline task difficulty does not directly determine pruning sensitivity.
Across all three datasets, degradation becomes substantially more
category-dependent as dense pruning becomes aggressive. In contrast, the MoE
model with REAP preserves relatively stable accuracy across task categories
over a much wider pruning range, with degradation emerging primarily at the
highest expert-pruning ratios.

\section{Operation-Level F1 Scores}
\label{app:f1}
\begin{figure*}[t]
\centering
\begin{subfigure}[t]{0.315\textwidth}\centering
\includegraphics[width=\linewidth]{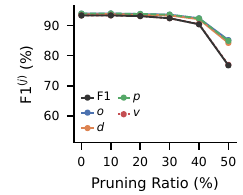}
\caption{Qwen3-4B, ShortGPT}
\end{subfigure}
\hfill
\begin{subfigure}[t]{0.315\textwidth}\centering
\includegraphics[width=\linewidth]{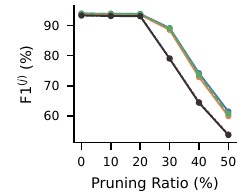}
\caption{Qwen3-4B, Angular}
\end{subfigure}
\hfill
\begin{subfigure}[t]{0.315\textwidth}\centering
\includegraphics[width=\linewidth]{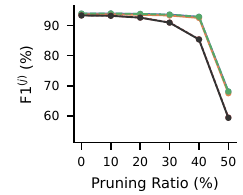}
\caption{Qwen3-4B, FLAP}
\end{subfigure}
\\[2pt]
\begin{subfigure}[t]{0.315\textwidth}\centering
\includegraphics[width=\linewidth]{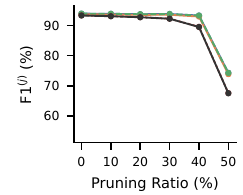}
\caption{Qwen3-4B, 2SSP}
\end{subfigure}
\hfill
\begin{subfigure}[t]{0.315\textwidth}\centering
\includegraphics[width=\linewidth]{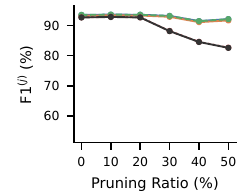}
\caption{Qwen3.5-4B, ShortGPT}
\end{subfigure}
\hfill
\begin{subfigure}[t]{0.315\textwidth}\centering
\includegraphics[width=\linewidth]{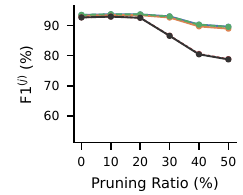}
\caption{Qwen3.5-4B, Angular}
\end{subfigure}
\\[2pt]
\begin{subfigure}[t]{0.315\textwidth}\centering
\includegraphics[width=\linewidth]{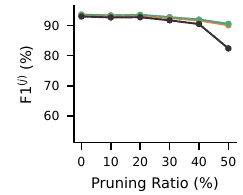}
\caption{Qwen3.5-9B, ShortGPT}
\end{subfigure}
\hfill
\begin{subfigure}[t]{0.315\textwidth}\centering
\includegraphics[width=\linewidth]{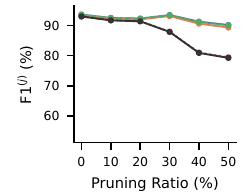}
\caption{Qwen3.5-9B, Angular}
\end{subfigure}
\hfill
\begin{subfigure}[t]{0.315\textwidth}\centering
\includegraphics[width=\linewidth]{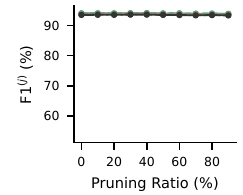}
\caption{Qwen3.6-35B-A3B, REAP}
\end{subfigure}
\caption{\textbf{HomeBench: operation-level $\mathrm{F1}$ of the action components.} $\mathrm{F1}$
together with the four action components $o,d,p,v$ against pruning ratio. Each subfigure is one LLM
and pruning method.}
\label{fig:panels_facets_homebench_f1}
\end{figure*}

Figures~\ref{fig:panels_facets_homebench_f1}-\ref{fig:panels_cats_greenv5_f1} repeat the
action-component and complexity-level analyses of Appendices~\ref{app:action-component}
and~\ref{app:complexity-level} with $\mathrm{F1}$ scores in place of $\mathrm{Acc}$. The action-component results are
Figures~\ref{fig:panels_facets_homebench_f1} (HomeBench), \ref{fig:panels_facets_har_f1} (HAR)
and~\ref{fig:panels_facets_greenv5_f1} (SHTC), and the complexity-level results are
Figures~\ref{fig:panels_cats_homebench_f1}, \ref{fig:panels_cats_har_f1}
and~\ref{fig:panels_cats_greenv5_f1}. We use the operation-level $\mathrm{F1}$ following \citet{li2025homebench}. Precision (P)  is
the number of operations the model predicts correctly divided by the number of operations it
predicts; recall (R) is the number of operations it predicts correctly divided by the number of
operations the user instruction actually requires, and $\mathrm{F1} = 2PR/(P+R)$.

$\mathrm{F1}$ does not change the conclusions, and the shape of the degradation is unchanged. Averaged over every LLM-pruning pair, the drop from $M_0$
to the most aggressive ratio decomposes almost identically under the two metrics: on SHTC
$d$ loses $28.7\%$ of $\mathrm{F1}$ against $20.9\%$ for $p$, $21.5\%$ for $v$ and $12.9\%$ for $o$, and on HAR $d$ loses $17.6\%$
points against $9.5\%$, $9.6\%$ and $7.8\%$.
Grounded target selection therefore remains the component that pruning erodes first, dense pruning
still shows a narrow safe region followed by a category-dependent collapse, and the MoE model under
REAP still stays flat over a far wider pruning range. 

\begin{figure*}
\centering
\begin{subfigure}[t]{0.315\textwidth}\centering
\includegraphics[width=\linewidth]{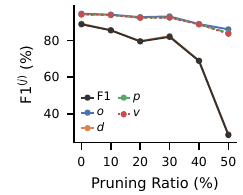}
\caption{Qwen3-4B, ShortGPT}
\end{subfigure}
\hfill
\begin{subfigure}[t]{0.315\textwidth}\centering
\includegraphics[width=\linewidth]{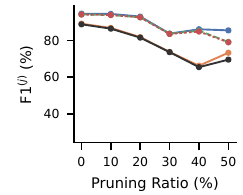}
\caption{Qwen3-4B, Angular}
\end{subfigure}
\hfill
\begin{subfigure}[t]{0.315\textwidth}\centering
\includegraphics[width=\linewidth]{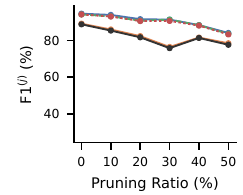}
\caption{Qwen3-4B, FLAP}
\end{subfigure}
\\[2pt]
\begin{subfigure}[t]{0.315\textwidth}\centering
\includegraphics[width=\linewidth]{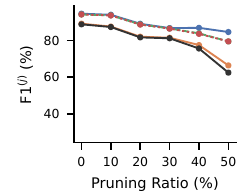}
\caption{Qwen3-4B, 2SSP}
\end{subfigure}
\hfill
\begin{subfigure}[t]{0.315\textwidth}\centering
\includegraphics[width=\linewidth]{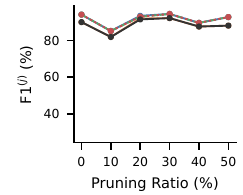}
\caption{Qwen3.5-4B, ShortGPT}
\end{subfigure}
\hfill
\begin{subfigure}[t]{0.315\textwidth}\centering
\includegraphics[width=\linewidth]{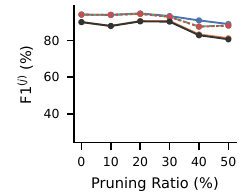}
\caption{Qwen3.5-4B, Angular}
\end{subfigure}
\\[2pt]
\begin{subfigure}[t]{0.315\textwidth}\centering
\includegraphics[width=\linewidth]{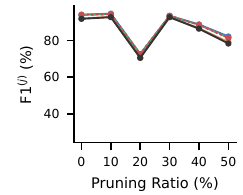}
\caption{Qwen3.5-9B, ShortGPT}
\end{subfigure}
\hfill
\begin{subfigure}[t]{0.315\textwidth}\centering
\includegraphics[width=\linewidth]{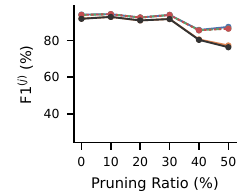}
\caption{Qwen3.5-9B, Angular}
\end{subfigure}
\hfill
\begin{subfigure}[t]{0.315\textwidth}\centering
\includegraphics[width=\linewidth]{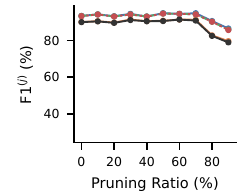}
\caption{Qwen3.6-35B-A3B, REAP}
\end{subfigure}
\caption{\textbf{HAR: operation-level $\mathrm{F1}$ of the action components.} $\mathrm{F1}$
together with the four action components $o,d,p,v$ against pruning ratio. Each subfigure is one LLM
and pruning method.}
\label{fig:panels_facets_har_f1}
\end{figure*}

\begin{figure*}[t]
\centering
\begin{subfigure}[t]{0.315\textwidth}\centering
\includegraphics[width=\linewidth]{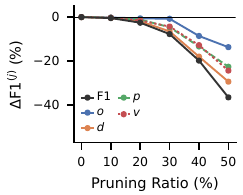}
\caption{Qwen3-4B, ShortGPT}
\end{subfigure}
\hfill
\begin{subfigure}[t]{0.315\textwidth}\centering
\includegraphics[width=\linewidth]{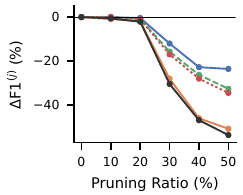}
\caption{Qwen3-4B, Angular}
\end{subfigure}
\hfill
\begin{subfigure}[t]{0.315\textwidth}\centering
\includegraphics[width=\linewidth]{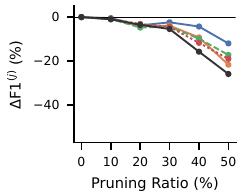}
\caption{Qwen3-4B, FLAP}
\end{subfigure}
\\[2pt]
\begin{subfigure}[t]{0.315\textwidth}\centering
\includegraphics[width=\linewidth]{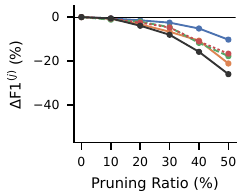}
\caption{Qwen3-4B, 2SSP}
\end{subfigure}
\hfill
\begin{subfigure}[t]{0.315\textwidth}\centering
\includegraphics[width=\linewidth]{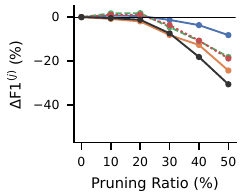}
\caption{Qwen3.5-4B, ShortGPT}
\end{subfigure}
\hfill
\begin{subfigure}[t]{0.315\textwidth}\centering
\includegraphics[width=\linewidth]{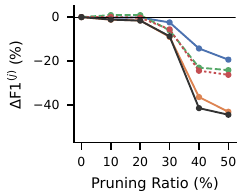}
\caption{Qwen3.5-4B, Angular}
\end{subfigure}
\\[2pt]
\begin{subfigure}[t]{0.315\textwidth}\centering
\includegraphics[width=\linewidth]{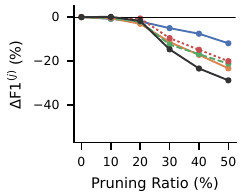}
\caption{Qwen3.5-9B, ShortGPT}
\end{subfigure}
\hfill
\begin{subfigure}[t]{0.315\textwidth}\centering
\includegraphics[width=\linewidth]{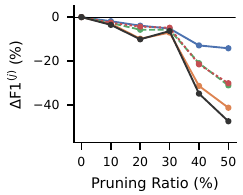}
\caption{Qwen3.5-9B, Angular}
\end{subfigure}
\hfill
\begin{subfigure}[t]{0.315\textwidth}\centering
\includegraphics[width=\linewidth]{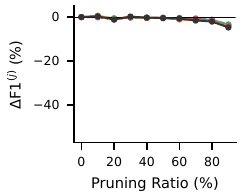}
\caption{Qwen3.6-35B-A3B, REAP}
\end{subfigure}
\caption{\textbf{SHTC: operation-level $\mathrm{F1}$ of the action components.}
$\Delta\mathrm{F1}$ relative to $M_0$ together with the four action components $o,d,p,v$ against
pruning ratio. Each subfigure is one LLM and pruning method.}
\label{fig:panels_facets_greenv5_f1}
\end{figure*}


\begin{figure*}[t]
\centering
\begin{subfigure}[t]{0.315\textwidth}\centering
\includegraphics[width=\linewidth]{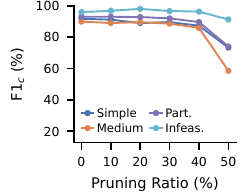}
\caption{Qwen3-4B, ShortGPT}
\end{subfigure}
\hfill
\begin{subfigure}[t]{0.315\textwidth}\centering
\includegraphics[width=\linewidth]{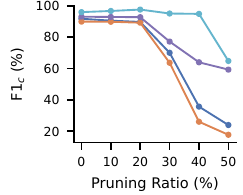}
\caption{Qwen3-4B, Angular}
\end{subfigure}
\hfill
\begin{subfigure}[t]{0.315\textwidth}\centering
\includegraphics[width=\linewidth]{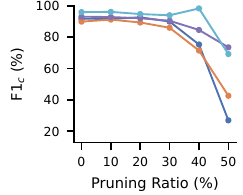}
\caption{Qwen3-4B, FLAP}
\end{subfigure}
\\[2pt]
\begin{subfigure}[t]{0.315\textwidth}\centering
\includegraphics[width=\linewidth]{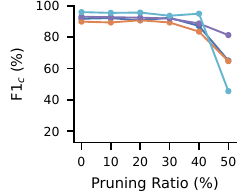}
\caption{Qwen3-4B, 2SSP}
\end{subfigure}
\hfill
\begin{subfigure}[t]{0.315\textwidth}\centering
\includegraphics[width=\linewidth]{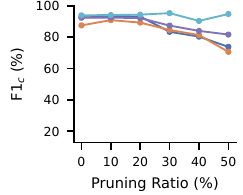}
\caption{Qwen3.5-4B, ShortGPT}
\end{subfigure}
\hfill
\begin{subfigure}[t]{0.315\textwidth}\centering
\includegraphics[width=\linewidth]{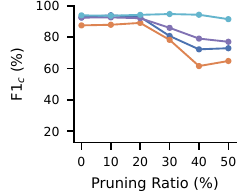}
\caption{Qwen3.5-4B, Angular}
\end{subfigure}
\\[2pt]
\begin{subfigure}[t]{0.315\textwidth}\centering
\includegraphics[width=\linewidth]{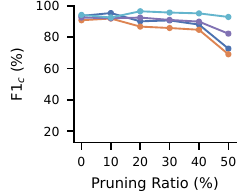}
\caption{Qwen3.5-9B, ShortGPT}
\end{subfigure}
\hfill
\begin{subfigure}[t]{0.315\textwidth}\centering
\includegraphics[width=\linewidth]{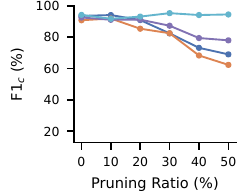}
\caption{Qwen3.5-9B, Angular}
\end{subfigure}
\hfill
\begin{subfigure}[t]{0.315\textwidth}\centering
\includegraphics[width=\linewidth]{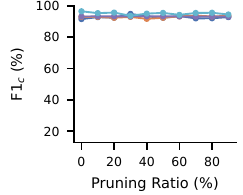}
\caption{Qwen3.6-35B-A3B, REAP}
\end{subfigure}
\caption{\textbf{HomeBench: operation-level $\mathrm{F1}$ of the task-complexity categories.}
$\mathrm{F1}_c$ for the complexity categories present in the source against pruning ratio. Each
subfigure is one LLM and pruning method.}
\label{fig:panels_cats_homebench_f1}
\end{figure*}

\begin{figure*}[t]
\centering
\begin{subfigure}[t]{0.315\textwidth}\centering
\includegraphics[width=\linewidth]{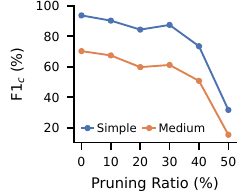}
\caption{Qwen3-4B, ShortGPT}
\end{subfigure}
\hfill
\begin{subfigure}[t]{0.315\textwidth}\centering
\includegraphics[width=\linewidth]{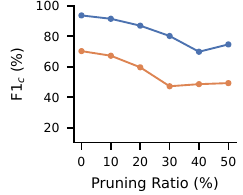}
\caption{Qwen3-4B, Angular}
\end{subfigure}
\hfill
\begin{subfigure}[t]{0.315\textwidth}\centering
\includegraphics[width=\linewidth]{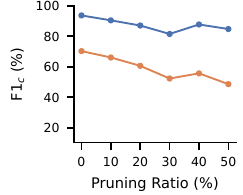}
\caption{Qwen3-4B, FLAP}
\end{subfigure}
\\[2pt]
\begin{subfigure}[t]{0.315\textwidth}\centering
\includegraphics[width=\linewidth]{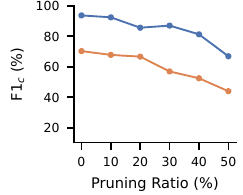}
\caption{Qwen3-4B, 2SSP}
\end{subfigure}
\hfill
\begin{subfigure}[t]{0.315\textwidth}\centering
\includegraphics[width=\linewidth]{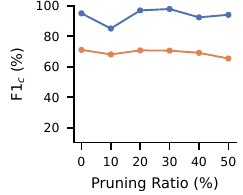}
\caption{Qwen3.5-4B, ShortGPT}
\end{subfigure}
\hfill
\begin{subfigure}[t]{0.315\textwidth}\centering
\includegraphics[width=\linewidth]{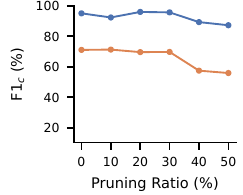}
\caption{Qwen3.5-4B, Angular}
\end{subfigure}
\\[2pt]
\begin{subfigure}[t]{0.315\textwidth}\centering
\includegraphics[width=\linewidth]{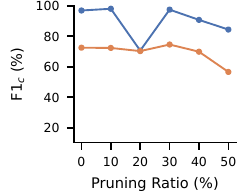}
\caption{Qwen3.5-9B, ShortGPT}
\end{subfigure}
\hfill
\begin{subfigure}[t]{0.315\textwidth}\centering
\includegraphics[width=\linewidth]{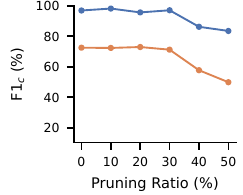}
\caption{Qwen3.5-9B, Angular}
\end{subfigure}
\hfill
\begin{subfigure}[t]{0.315\textwidth}\centering
\includegraphics[width=\linewidth]{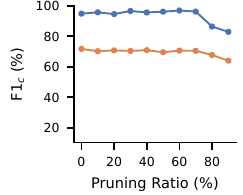}
\caption{Qwen3.6-35B-A3B, REAP}
\end{subfigure}
\caption{\textbf{HAR: operation-level $\mathrm{F1}$ of the task-complexity categories.}
$\mathrm{F1}_c$ for the complexity categories present in the source against pruning ratio. Each
subfigure is one LLM and pruning method.}
\label{fig:panels_cats_har_f1}
\end{figure*}

\begin{figure*}[t]
\centering
\begin{subfigure}[t]{0.315\textwidth}\centering
\includegraphics[width=\linewidth]{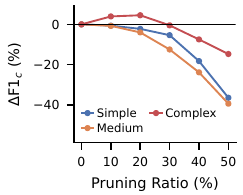}
\caption{Qwen3-4B, ShortGPT}
\end{subfigure}
\hfill
\begin{subfigure}[t]{0.315\textwidth}\centering
\includegraphics[width=\linewidth]{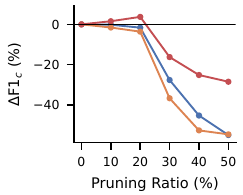}
\caption{Qwen3-4B, Angular}
\end{subfigure}
\hfill
\begin{subfigure}[t]{0.315\textwidth}\centering
\includegraphics[width=\linewidth]{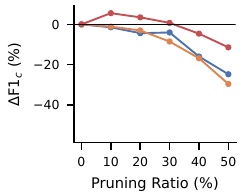}
\caption{Qwen3-4B, FLAP}
\end{subfigure}
\\[2pt]
\begin{subfigure}[t]{0.315\textwidth}\centering
\includegraphics[width=\linewidth]{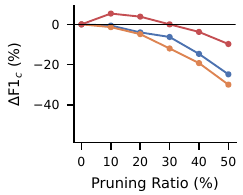}
\caption{Qwen3-4B, 2SSP}
\end{subfigure}
\hfill
\begin{subfigure}[t]{0.315\textwidth}\centering
\includegraphics[width=\linewidth]{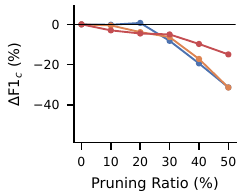}
\caption{Qwen3.5-4B, ShortGPT}
\end{subfigure}
\hfill
\begin{subfigure}[t]{0.315\textwidth}\centering
\includegraphics[width=\linewidth]{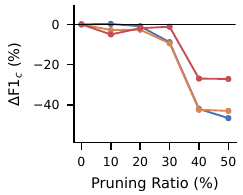}
\caption{Qwen3.5-4B, Angular}
\end{subfigure}
\\[2pt]
\begin{subfigure}[t]{0.315\textwidth}\centering
\includegraphics[width=\linewidth]{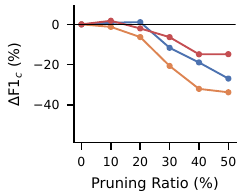}
\caption{Qwen3.5-9B, ShortGPT}
\end{subfigure}
\hfill
\begin{subfigure}[t]{0.315\textwidth}\centering
\includegraphics[width=\linewidth]{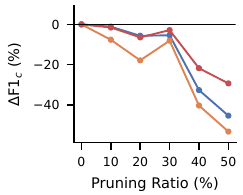}
\caption{Qwen3.5-9B, Angular}
\end{subfigure}
\hfill
\begin{subfigure}[t]{0.315\textwidth}\centering
\includegraphics[width=\linewidth]{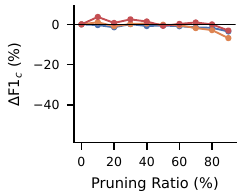}
\caption{Qwen3.6-35B-A3B, REAP}
\end{subfigure}
\caption{\textbf{SHTC: operation-level $\mathrm{F1}$ of the task-complexity categories.}
$\Delta\mathrm{F1}_c$ relative to $M_0$ for the complexity categories present in the source against
pruning ratio. Each subfigure is one LLM and pruning method.}
\label{fig:panels_cats_greenv5_f1}
\end{figure*}

\section{Robustness Analyses}\label{app_sec_sob}
\subsection{Robustness Across the Pruning Ratios}
To study the behavioral robustness across pruning ratios, we use the results at each REAP ratio for Qwen3.6-35B-A3B. For each
instance and metric, we classify the outcome across the ten checkpoints from
$0\%$ to $90\%$ as always correct, always wrong, or flipping across ratios. Thus, this analysis measures whether the same
instances remain correct throughout the pruning sweep, rather than variation
across random seeds.

\begin{figure*}
\centering
\begin{subfigure}[t]{0.315\textwidth}
\centering
\includegraphics[width=\linewidth]{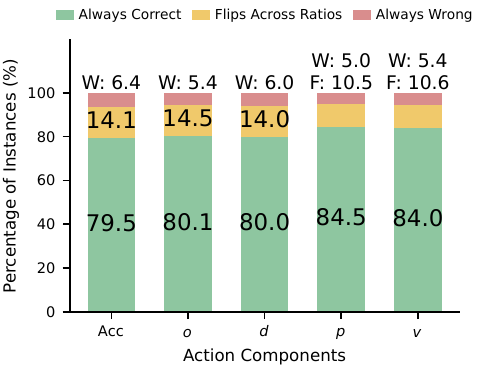}
\caption{HomeBench: Action Components}
\end{subfigure}
\hfill
\begin{subfigure}[t]{0.315\textwidth}
\centering
\includegraphics[width=\linewidth]{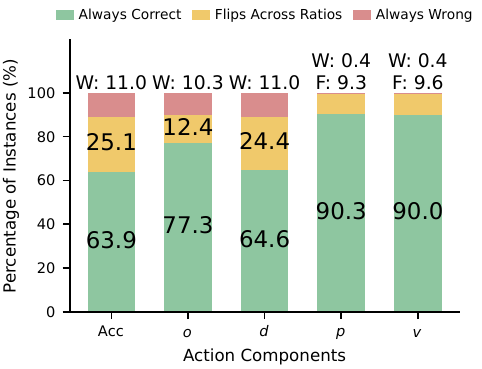}
\caption{HAR: Action Components}
\end{subfigure}
\hfill
\begin{subfigure}[t]{0.315\textwidth}
\centering
\includegraphics[width=\linewidth]{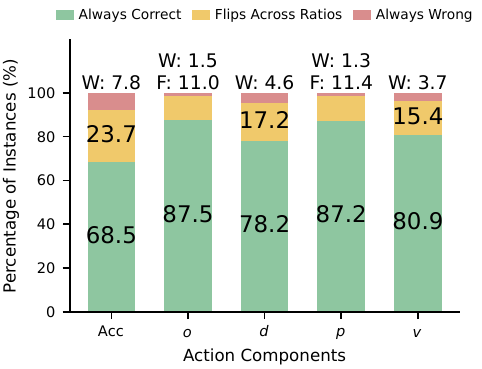}
\caption{SHTC: Action Components}
\end{subfigure}
\\[3pt]
\begin{subfigure}[t]{0.315\textwidth}
\centering
\includegraphics[width=\linewidth]{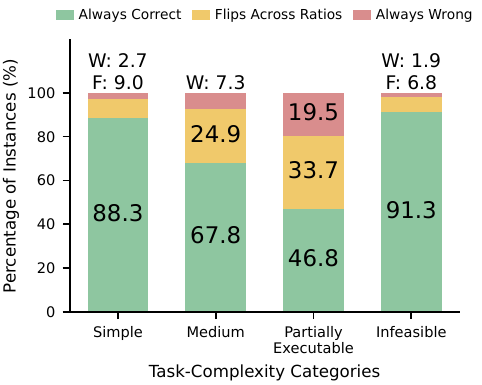}
\caption{HomeBench: Task-Complexity Categories}
\end{subfigure}
\hfill
\begin{subfigure}[t]{0.315\textwidth}
\centering
\includegraphics[width=\linewidth]{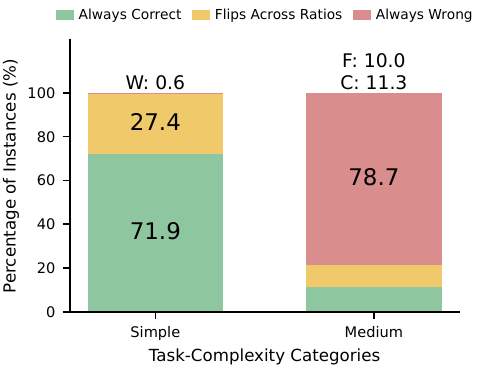}
\caption{HAR: Task-Complexity Categories}
\end{subfigure}
\hfill
\begin{subfigure}[t]{0.315\textwidth}
\centering
\includegraphics[width=\linewidth]{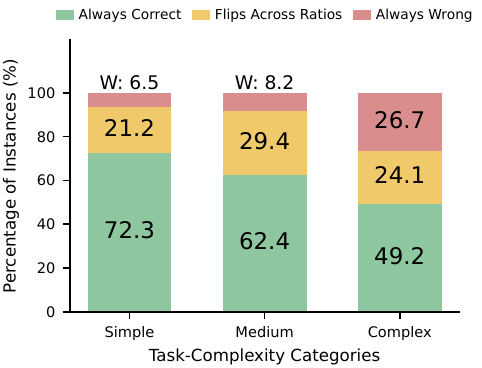}
\caption{SHTC: Task-Complexity Categories}
\end{subfigure}
\caption{Qwen3.6 MoE robustness across REAP pruning ratios by action component and task-complexity level. Labels prefixed by C, F, or W
above a bar denote Always Correct, Flips Across Ratios, or Always Wrong.}
\label{fig:greedy-moe-ratio-stability}
\end{figure*}

Figure~\ref{fig:greedy-moe-ratio-stability} shows that robust aggregate
accuracy does not imply stable instance-level behavior. On HomeBench,
$14.1\%$ of exact-match outcomes change somewhere along the pruning, rising to
$33.7\%$ for Partially Executable and $24.9\%$ for Medium requests, despite
the nearly flat aggregate curve. HAR has a larger exact-match flipping band
($25.1\%$), concentrated in device grounding ($24.4\%$) and Simple requests
($27.4\%$); its Medium band is instead mostly always wrong ($78.7\%$).
On SHTC, Medium is again the least stable category ($29.4\%$ flipping), while
operation selection is the most stable action component ($11.0\%$). These
results distinguish a stable mean from stable predictions: expert pruning can
preserve aggregate accuracy while changing which requests succeed.

\subsection{Robustness Across Seeds}
\label{app:stability}

We measure behavioral robustness using three reruns with the same weights and
sampling settings (i.e., temperature is 0.7) for unpruned Qwen3.6-35B-A3B. For a given instance and metric,
an outcome is always correct if all three runs succeed, always
wrong if all three fail, and flipping if correctness changes across
seeds. The flipping share therefore measures cross-seed instability, whereas
the always-wrong share measures a seed-invariant failure core.

\begin{figure*}
\centering
\begin{subfigure}[t]{0.315\textwidth}
\centering
\includegraphics[width=\linewidth]{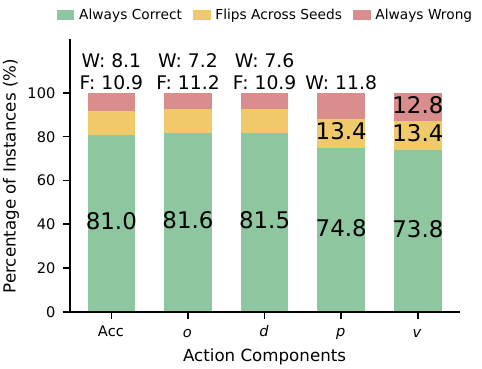}
\caption{HomeBench: Action Components}
\end{subfigure}
\hfill
\begin{subfigure}[t]{0.315\textwidth}
\centering
\includegraphics[width=\linewidth]{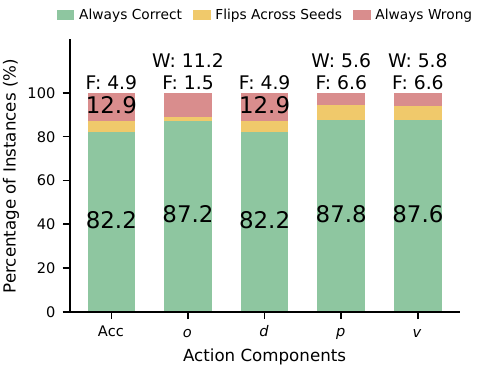}
\caption{HAR: Action Components}
\end{subfigure}
\hfill
\begin{subfigure}[t]{0.315\textwidth}
\centering
\includegraphics[width=\linewidth]{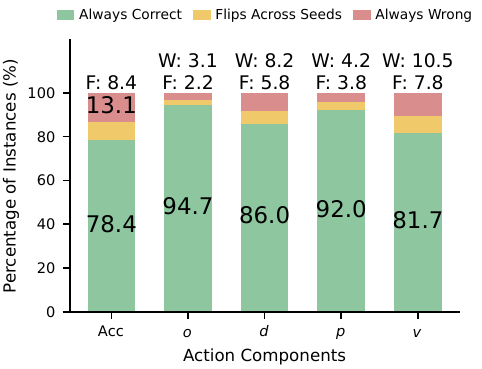}
\caption{SHTC: Action Components}
\end{subfigure}
\\[3pt]
\begin{subfigure}[t]{0.315\textwidth}
\centering
\includegraphics[width=\linewidth]{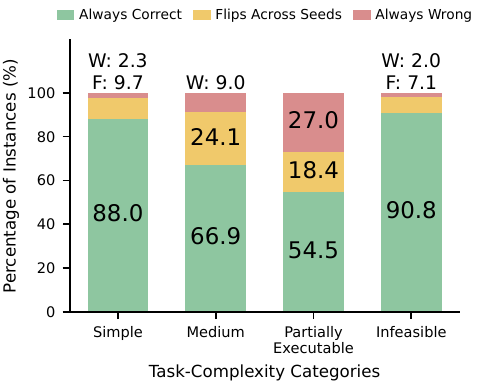}
\caption{HomeBench: Task-Complexity Categories}
\end{subfigure}
\hfill
\begin{subfigure}[t]{0.315\textwidth}
\centering
\includegraphics[width=\linewidth]{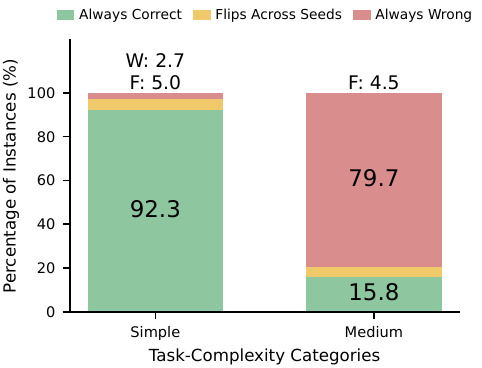}
\caption{HAR: Task-Complexity Categories}
\end{subfigure}
\hfill
\begin{subfigure}[t]{0.315\textwidth}
\centering
\includegraphics[width=\linewidth]{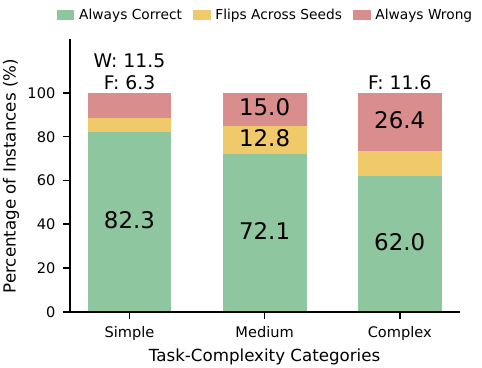}
\caption{SHTC: Task-Complexity Categories}
\end{subfigure}
\caption{Qwen3.6 MoE robustness across seeds by action component and task-complexity level. Labels prefixed by C, F, or W
above a bar denote Always Correct, Flips Across Seeds, or Always Wrong.}
\label{fig:seed-stability-reference}
\end{figure*}

Figure~\ref{fig:seed-stability-reference} shows that task difficulty and
cross-seed instability are not equivalent. HomeBench Medium has the largest
borderline band ($24.1\%$ flipping), followed by Partially Executable
($18.4\%$), while Infeasible is comparatively stable ($7.1\%$). In contrast,
HAR Medium flips on only $4.5\%$ of instances but is always wrong on $79.7\%$:
it is chronically difficult rather than seed-sensitive. On SHTC, Medium and
Complex flip more often than Simple ($12.8\%$ and $11.6\%$ versus $6.3\%$).

\end{document}